\documentclass{bmvc2k}

\title{Illuminated Decision Trees with Lucid}

\addauthor{David Mott}{http://research.ibm.com/labs/uk/}{1}
\addauthor{Richard Tomsett}{http://research.ibm.com/labs/uk/}{1}

\addinstitution{
 IBM Emerging Technology\\
 IBM Research\\
 Hursley, UK
}

\runninghead{Mott \& Tomsett}{Illuminated Decision Trees with Lucid}

\begin{document}

\maketitle

\begin{abstract}
The Lucid methods described by Olah et al.~\cite{olah2018the} provide a way to inspect the inner workings of neural networks trained on image classification tasks using feature visualization. Such methods have generally been applied to networks trained on visually rich, large-scale image datasets like ImageNet, which enables them to produce enticing feature visualizations. To investigate these methods further, we applied them to classifiers trained to perform the much simpler (in terms of dataset size and visual richness), yet challenging task of distinguishing between different kinds of white blood cell from microscope images. Such a task makes generating useful feature visualizations difficult, as the discriminative features are inherently hard to identify and interpret. We address this by presenting the ``Illuminated Decision Tree'' approach, in which we use a neural network trained on the task as a feature extractor, then learn a decision tree based on these features, and provide Lucid visualizations for each node in the tree. We demonstrate our approach with several examples, showing how this approach could be useful both in model development and debugging, and when explaining model outputs to non-experts.
\end{abstract}

\section{Introduction}
\label{sec:intro}
Visualizing the features learned by neural networks (NNs) has been suggested as a powerful method for helping to understand what NNs have learned~\cite{zeiler14, YosinskiCNFL15, Nguyen16, olah2017feature, olah2018the}. Recent impressive work by Olah et al. illustrated the strength of their feature visualization methods by examining a modern image classifier trained to recognise 1000 different object classes~\cite{olah2017feature,olah2018the}. Their results show how different channels and neurons within a NN learn to respond to identifiable/semantic sub-features contained within the training data. While such visualizations are compelling, it is unclear how useful they might be when applied to smaller, simpler datasets, or to image classification problems that are more domain-specific or abstract.

We investigated the Lucid feature visualization methods by exploring a task available through the Kaggle machine learning challenge web-site: discriminating between different types of white blood cell in microscope images~\cite{kaggle}. This dataset is much simpler than, for example, the standard 1000--class ImageNet~\cite{ILSVRC15}, as there are orders of magnitude fewer classes and the images are much more homogeneous. However, the machine learning task is challenging, as the discriminating features between different classes are non-obvious and subtle, sometimes even to experts with relevant domain knowledge. This makes it an interesting problem on which to apply Lucid visualization methods.

In the remainder of the paper, we will describe the cell classification problem, show some results of the Lucid visualizations on our trained models, and show how we can use these visualizations to construct more interpretable classifiers that we call ``Illuminated Decision Trees.''

\begin{figure*}
\begin{center}
\includegraphics[height=5cm]{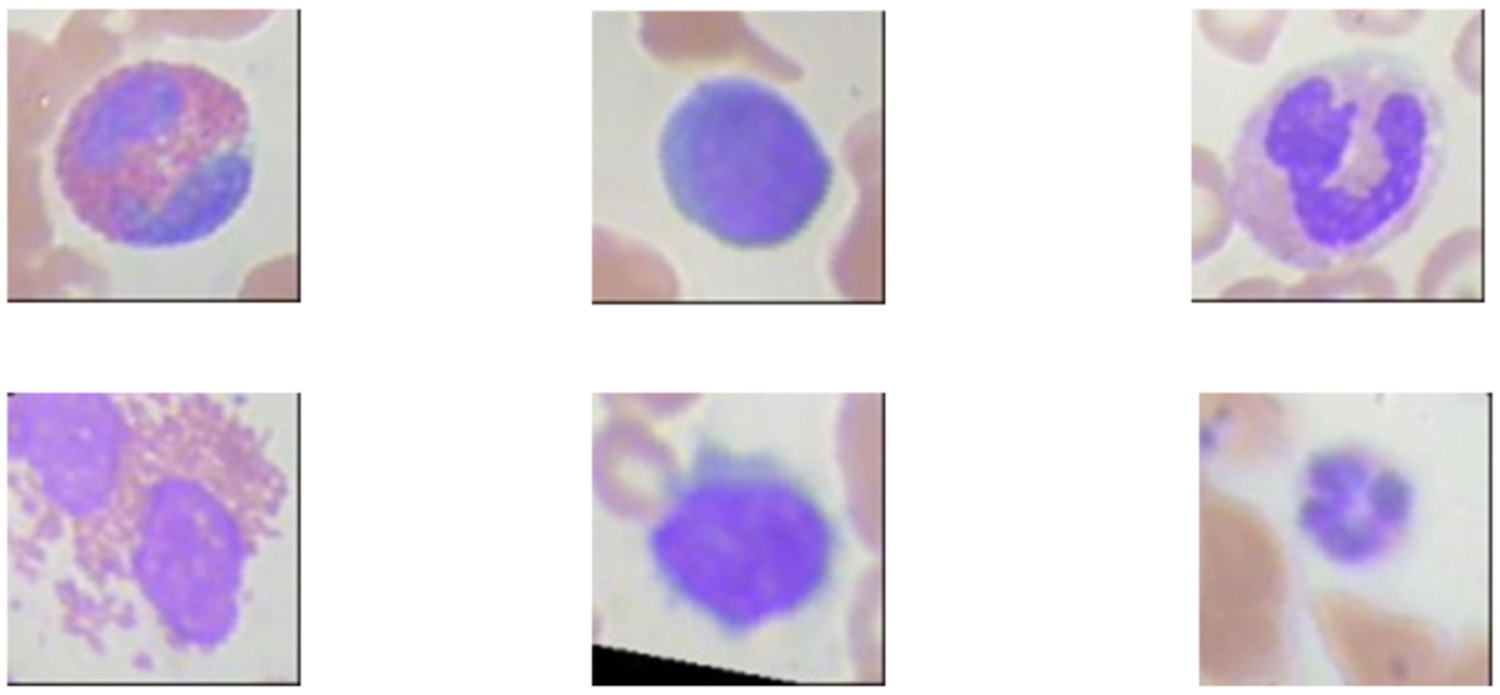}
\end{center}
   \caption{Example white blood cells in the Kaggle dataset~\cite{kaggle}, extracted as described in the text. The left column shows, eosinophils; the central column shows lymphocytes, the right column shows neutrophils}
\label{fig:short}
\end{figure*}

\section{Data}
\label{sec:data}
Kaggle provides two datasets of white blood cells, with the first, ``dataset-master'' containing a fewer images than the second ``dataset-master2''~\cite{kaggle}. Dataset-master2 has four different types of cell: eosinophil, lymphocyte, monocyte and neutrophil, with the following number of images in each class --- eosinophil: 3133, lymphocyte: 3109, monocyte: 3102, neutrophil: 3171. The white blood cells are quite small compared to the whole image, with extensive background material such as red blood cells and black edges. Therefore code was written to extract the cells (using a simple colour area detection followed by the application of a bounding box to the detected area) into a JPEG image of 100$\times$100 pixels with three 8--bit colour channels. We chose to focus on two problems of differing difficulty: an easy problem of distinguishing between lymphocytes and eosinophils (task ``LE''), and a hard problem of distinguishing between eosinophils and neutrophils (task ``EN''). Example blood cell images are shown in Figure 1. 

\section{Feature Visualization}
\label{sec:featureviz}
Figure 2 shows features learned at layers 1 and 4 of a 4-layer CNN trained on task LE. The Lucid feature visualization method focuses on a single feature at a time and aims to create an image that ``most activates'' this feature. Such an image is created by starting with a random image, running it through the classifier, working out the direction (gradients) in which to adjust the image to make a more accurate classification (thus reducing the loss between the prediction and the actual). Each layer contains multiple features and for each one we create the image to maximise the response of that feature as averaged over all spatial positions.

\begin{figure*}
\begin{center}
\includegraphics[height=6cm]{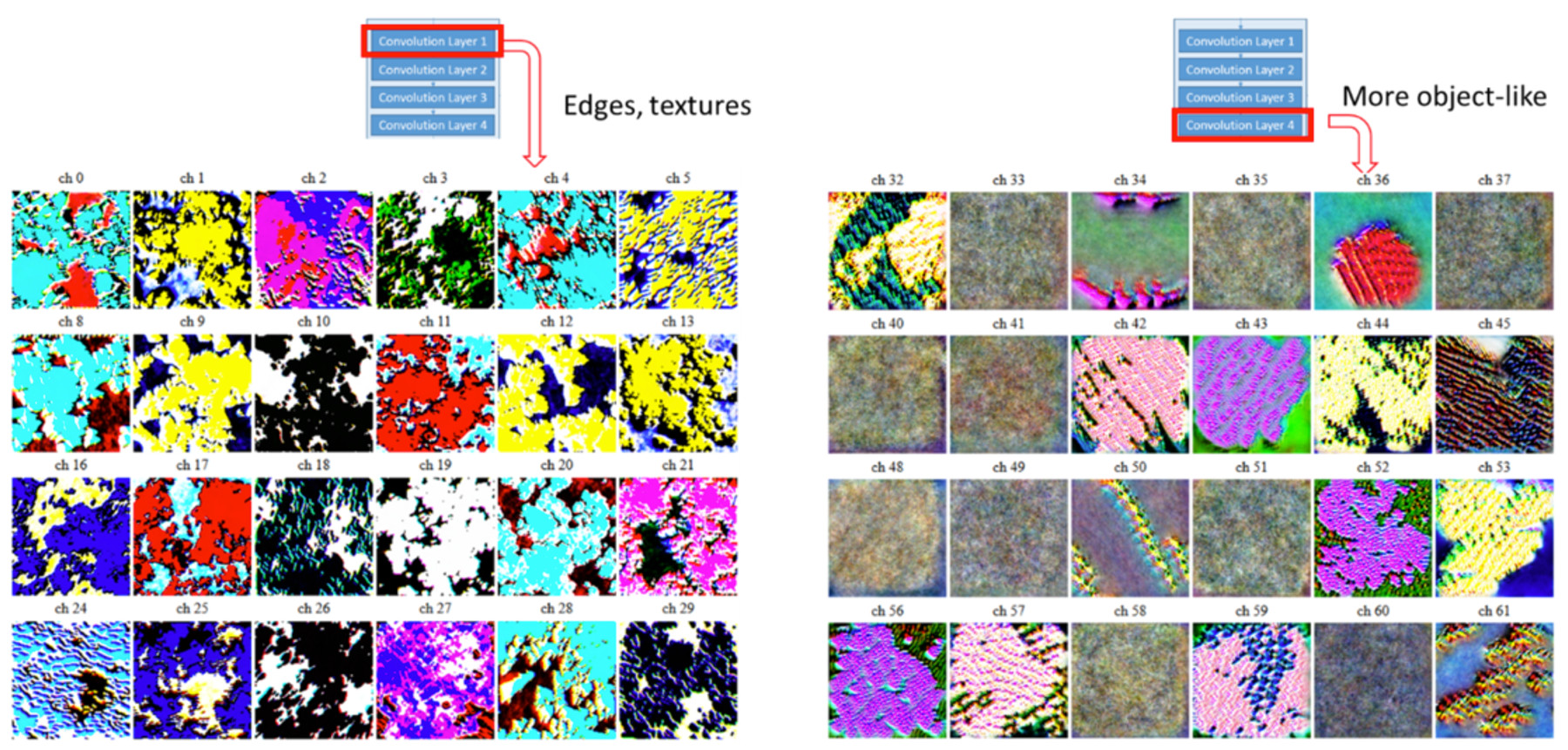}
\end{center}
   \caption{Visualizing the features learned by a 4-layer CNN on task LE}
\label{fig:short}
\end{figure*}

In Figure 2, features on two layers are visualised. On the left is the first 24 features of the shallowest layer 1 (closest to the blood cell image) and shows textures and edges. The colour is related to the colour that the features match on, but these visualisations tend to be exaggerated for colour. On the right are some of the features of the deepest layer 4, and this shows images that are more like objects or structures, since this an aggregation, or abstraction, of the layers above (1-3). Some of the features are grey, suggesting that they did not contribute anything to the classification (we noted that more accurate classifiers seem in some cases to have more ``active'' features).

To visualise how the features match against an image, ``best channel images'' may be created. Such images may provide confirmation that the Lucid visualisations are sensible and informative. An example is shown in Figure 3: here, a particular image (of an eosinophil) has been run through the CNN and the best matched feature at each point on a given layer has been calculated and turned into a 2D map with the same topology as the original image. This ``best channel image'' is taken at layer 4M which is a 10$\times$10 matrix, with 128 possible features at each point. At each point in the 10$\times$10 matrix we display the visualisation of the best (most activated) feature at that point, and a larger square represents a stronger activation. At the bottom left is displayed the channel number (feature number) used in the main diagram. This reveals that channel 44 is largely activated by the background of the image rather than by features of the blood cell itself.

\begin{figure*}
\begin{center}
\includegraphics[height=7cm]{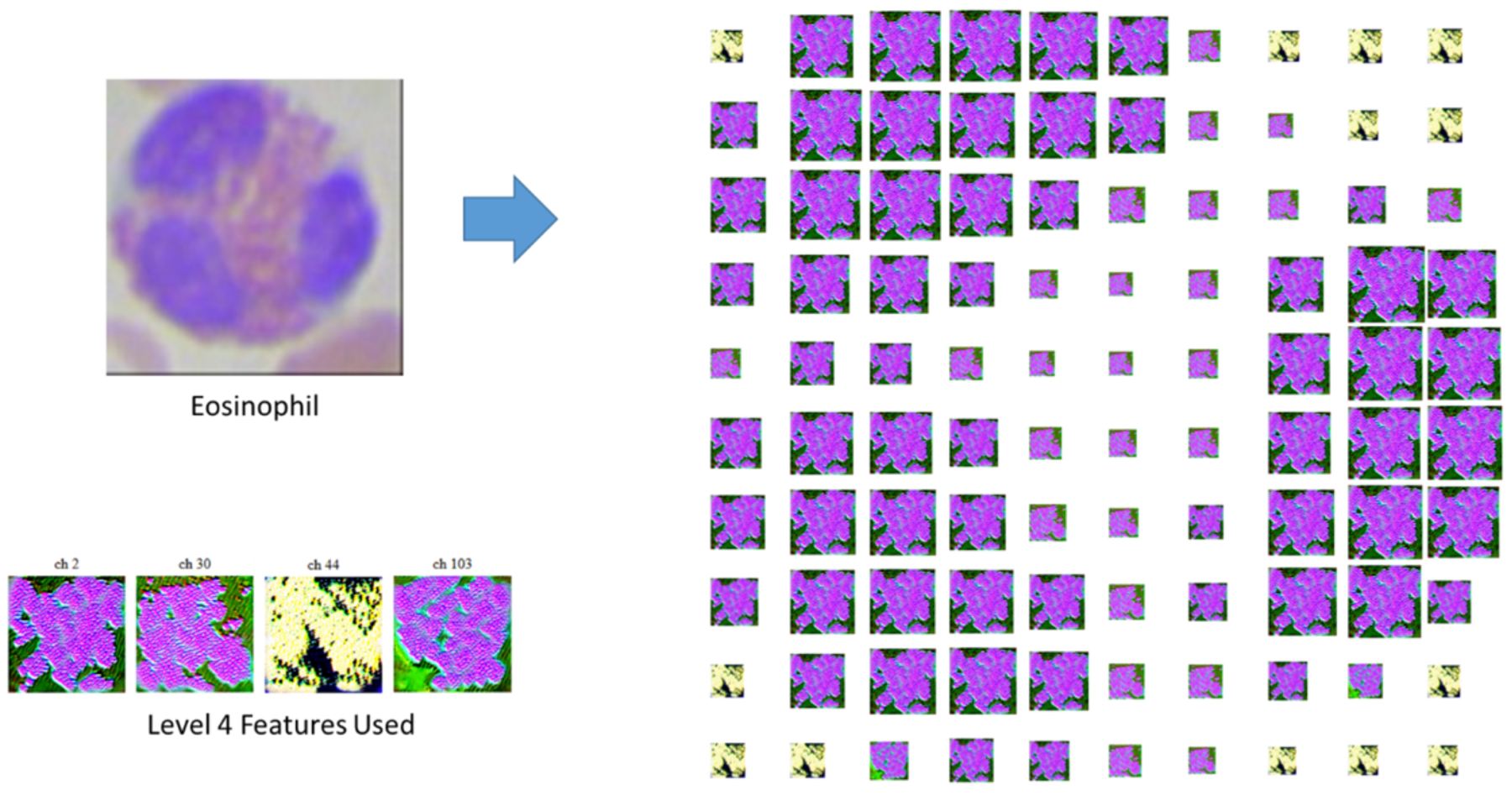}
\end{center}
   \caption{Strongest features in an image (4-layer CNN, task LE}
\label{fig:short}
\end{figure*}

\section{Illuminated Decision Trees}
\label{sec:illuminated}
We developed the concept of an Illuminated Decision Tree in an effort to create a more easily interpretable model that combined the feature extraction power of a CNN with the easily followed decision pathways learned by a decision tree classifier (note that this is only possible as we are studying a problem with a small number of classes --- if we were to build illuminated decision trees for ImageNet, the number of branches needed to distinguish the 1000 classes would render the trees almost totally uninterpretable). The CNN feature extractor is used to create a training set of <feature vector, target class> pairs by running each image through the classifier. This training set is used to construct the decision tree. Each node on the tree has a criterion, which is the comparison of a feature to a threshold: if the feature is greater than the threshold then the left-hand branch is taken. Previous work has 

We can then use the feature visualizations to illustrate the decision tree, providing some level of interpretability for the decision model. The example tree shown in figure 4 was learned using features from a 4-layer CNN, trained on the simple binary classification task. Here the lymphocytes and eosinophils are mostly discriminated via the first feature 6\_5\_9, i.e. channel 9 on location 6,5 of the output map. To a human, this 2-class problem appears to be to discriminate ``pink things'' (eosinophils) from ``blue things'' (lymphocytes), so it is encouraging that the discriminating test in the decision tree is based upon a pink feature. (There are further discriminations that mostly seek to deal with smaller numbers of outliers). It should be noted that the features at the next layer down are yellow and are at locations 5,7 and 5,0 on the output map. The accuracy of this decision tree is 0.88, while the corresponding CNN classifier obtained an accuracy of 0.93. Allowing the decision tree to grow deeper improves its accuracy at the cost of making the tree more difficult to interpret.

\begin{figure*}
\begin{center}
\includegraphics[height=5cm]{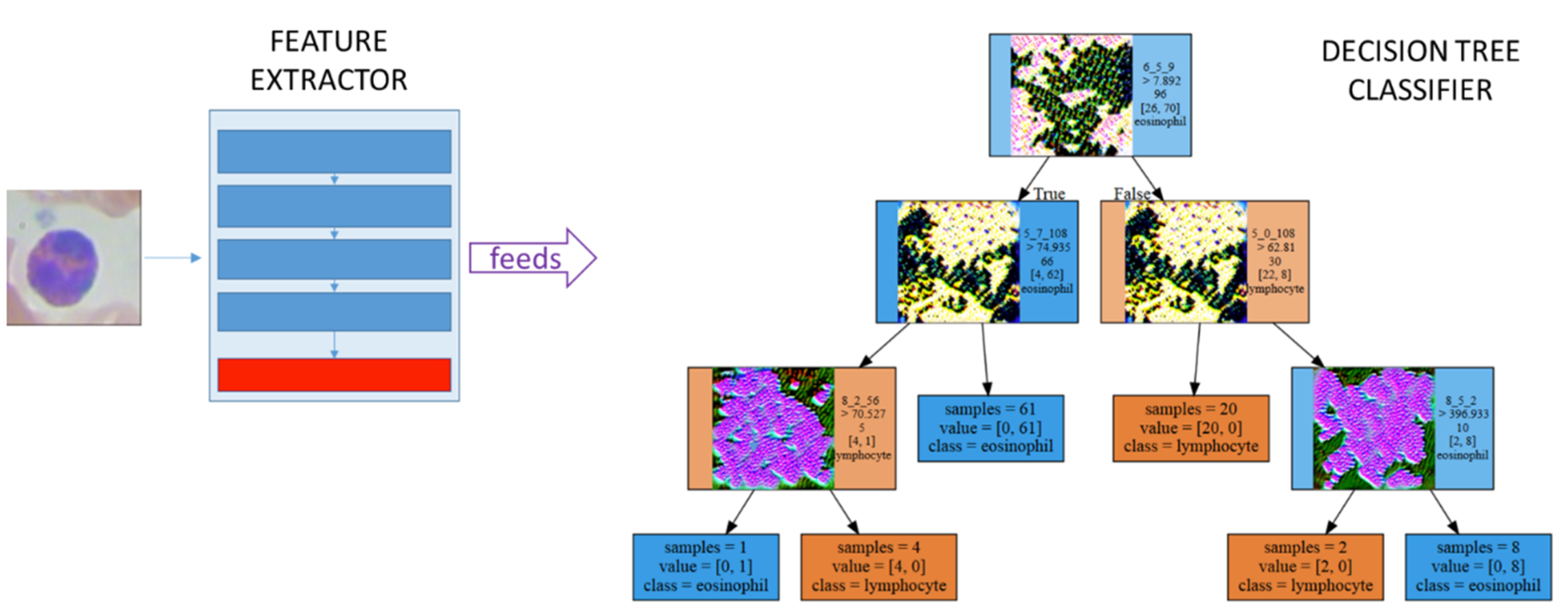}
\end{center}
   \caption{Creating an illuminated decision tree using features learned by a CNN. In this case, a 4-layer CNN has been trained on the 2-class problem.}
\label{fig:short}
\end{figure*}

The EN task is significantly more difficult than the LE task because eosinophils and neutrophils are difficult to distinguish from each other in this dataset. To obtain reasonable accuracy, we trained a 6-layer CNN with a larger set of images than previously used for task LE. The resulting illuminated decision tree is shown in figure 5; this obtains an accuracy of 0.82. Here the top node sends 57\% of the eosinophils to the left and 88\% of the neutrophils to the right. The eosinophils have one leaf node that contains a large number of images, with the remainder of the eosinophils and all of the neutrophils spread thinly across the other leaf nodes. More importantly some of the leaf nodes fail to discriminate their sets. Again this situation could be improved by a deeper tree, but the general characteristics of this tree seem to be engendered by the difficultly of the classification task.

In regard to the features, pale pink features seem to indicate eosinophil (e.g. the node on the middle left with feature 0\_0\_20 sends most of the eosinophils to the left and most of the neutrophils to the right), whereas grainy blue-pink features seem to indicate neutrophil. Some indication of the nature of the features can be gained by considering what they fail to match: for example, the top node (a pale pink feature) sends most of the neutrophils to the right, i.e. they fail to match the pale pink and absence of pale pink is indicative of neutrophils. However, this feature also fails to match some eosinophils, so it is not totally discriminating; nevertheless whenever there is a pale pink component of a feature (23,20,41,53) the neutrophils almost invariably go to the right (i.e they fail to match).

\begin{figure*}
\begin{center}
\includegraphics[height=18cm]{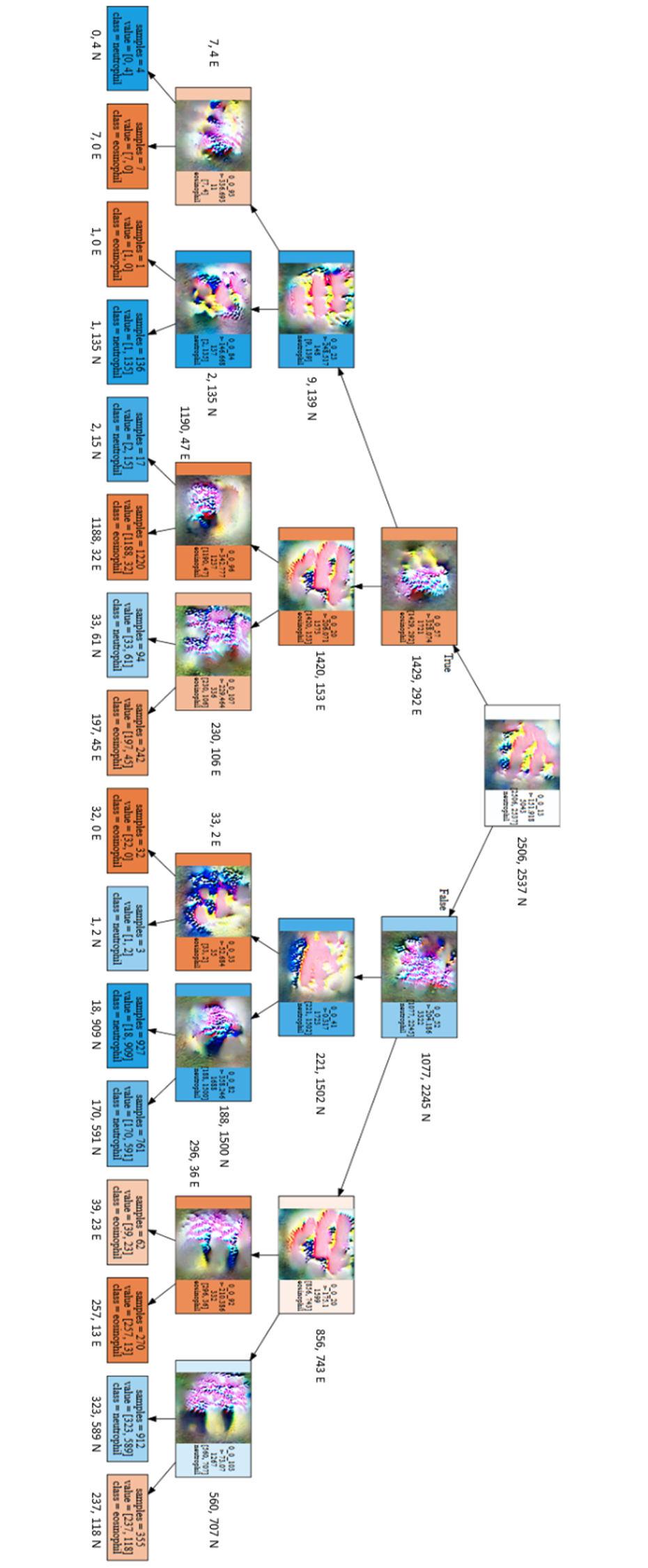}
\end{center}
   \caption{The learned illuminated decision tree for the 3-class problem, based on features learned by a 6-layer CNN. The feature visualisations appear much richer than the blob-like features shown in figure 4, illustrating that a more complex visual model has been learned at the expense of interpretability of each feature.}
\label{fig:short}
\end{figure*}

\section{Discussion}
\label{sec:discussion}
We have presented some examples of how the Lucid feature visualisation methods can be used to investigate image classifiers trained on visually simple, but difficult classification tasks. The feature visualisations follow a similar pattern to those of more complex ImageNet classifiers, with lower layers responding to simple features and higher layers responding to more high-level structure. However, the white blood-cell classification task, with its relatively small quantity of data, small number of classes, and visually similar classes, produces feature visualizations that are sometimes difficult to understand other than in terms of their overall colour. We showed how constructing illuminated decision trees based on these features could both help with the interpretation of the visualizations, and create more understandable decision tree classifiers without a large reduction in accuracy compared with the black box CNN. This helped identify biases in the training data (classifiers focusing on areas of background rather than on the cells, for example). The illuminated decision tree approach provides options for dealing with such biases if we cannot retrain the CNN features with better data: we can keep the feature but be aware of its effects on classification due to the visualization of the decision tree, or we can exclude the feature from the tree learning. This provides the modeler with additional flexibility when exploring their models and data.

\section{Acknowledgement}
\label{sec:acknowledgement}
This research was sponsored by the U.S. Army Research Laboratory and the UK Ministry of Defence under Agreement Number W911NF-16-3-0001. The views and conclusions contained in this document are those of the authors and should not be interpreted as representing the official policies, either expressed or implied, of the U.S. Army Research Laboratory, the U.S. Government, the UK Ministry of Defence or the UK Government. The U.S. and UK Governments are authorized to reproduce and distribute reprints for Government purposes notwithstanding any copy-right notation hereon.

\bibliography{bmvc_final}
\end{document}